\documentclass[final]{clv2025}

\jvol{vv}
\jnum{nn}
\jyear{2025}

\dochead{Squib} 

\pageonefooter{Action Editor: Michael White. Submission received: 26 April 2026; revised version received: 21 July 2026; accepted for publication: 19 August 2026.}

\usepackage{amsmath}
\usepackage{booktabs}
\usepackage[dvipsnames]{xcolor}
\usepackage{subcaption}

\runningtitle{Papers Without Code}
\runningauthor{Meyer \& Roth}

\begin{document}

\title{Papers Without Code: Availability of GitHub Repositories Linked in *CL Publications}

\author{Selina Meyer\thanks{Corresponding author}$^{}$, Michael Roth$^{}$}

\affilblock{
    \affil{University of Technology Nuremberg\\\quad \email{selina.meyer@utn.de}\\\quad \email{michael.roth@utn.de}}
}

\maketitle

\begin{abstract}
Source code and data published at computational linguistics \textit{(*CL)} venues are increasingly being shared via GitHub. While this generally is a favourable development for the accessibility and potential reusability of research artifacts in natural language processing (NLP), the long-term availability of such repositories has not been evaluated. In this squib, we discuss the availability of 
repositories linked in papers published in the Computational Linguistics (CL) journal as well as at ACL and its co-located events over the past ten years. Contrary to our expectations, we find that GitHub repositories linked in more recent ACL publications are unavailable at similar rates as in older publications, in parts due to an increase in empty and placeholder repositories. Similar trends hold for other *CL venues, but not for platforms other than GitHub. 
\end{abstract}

\section{Introduction}

Publicly sharing research artifacts such as code and data is treated as a key requirement for reproducibility and impactful research across disciplines \cite{magnusson-etal-2023-reproducibility, celi2019plos, SHAMIR201354, eglen2017toward, wieling-etal-2018-squib} and as a component of good scientific practice in light of frameworks such as the FAIR principles \cite{wilkinson2016fair} and code of conducts of national science funding organizations \cite{nsf_pappg_ch11D4_2025, deutsche_forschungsgemeinschaft_2025_14281892, ukri_making_your_research_data_open_2025}, some of which mandate long-term storage of a minimum of ten years. In the area of natural language processing (NLP), such artifacts are predominantly shared on GitHub \cite{mieskes-etal-2019-community}, but little is known about their long-term availability.

Here, we evaluate the availability of such repositories by extracting GitHub links from papers published in the Computational Linguistics (CL) journal as well as the ACL conference and its co-located events (hereafter collectively referred to as ``ACL papers'') over the past ten years and checking their availability using the GitHub API. Results show that even among ACL papers published in 2025, a substantial
number link to unavailable repositories, many of which are associated with the papers' own research artifacts. A comparison with other *CL venues and online repositories shows that (a) this issue is not exclusive to ACL and (b) it most commonly occurs with GitHub. 
Building on these observations, we provide recommendations for authors, reviewers, organizers and editors  to improve long-term code and data availability, along with a low-resource Python script integrated in aclpubcheck for automatically checking GitHub repository availability in camera-ready papers.\footnote{Aclpubcheck extension: \href{https://zenodo.org/records/19497034}{10.5281/zenodo.19497034}; Paper Code: \href{https://zenodo.org/records/19591044}{10.5281/zenodo.19591044}}

\begin{figure}
    \centering
    \includegraphics[width=0.65\linewidth]{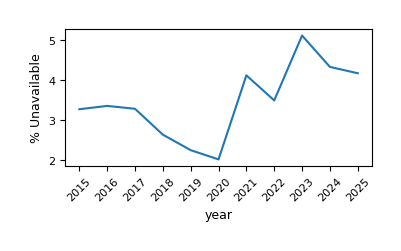}
    \caption{Share of unavailable GitHub repositories linked in ACL papers. While older links might reasonably decay, the consistent unavailability in recent years is concerning.}
    \label{fig:share_unavailable}
\end{figure}

\section{Related Work}
Reproducibility has gained increasing importance in NLP in recent years, largely as a consequence of the reproducibility crisis permeating many scientific disciplines, often due to unavailability of essential information, code, or data \cite{miyakawa2020no, belz-etal-2023-non, hutson2018artificial}. Dedicated tutorials and shared tasks \cite{lucic-etal-2022-towards, belz-etal-2025-2025, belz-thomson-2024-2024}, reproducibility checklists at *CL venues \cite{dodge-etal-2019-show} and the piloting of a reproducibility track at NAACL 2022 \cite{naacl2022_reproducibility_track_2022} represent efforts towards more open and transparent science, while others work towards streamlining the definition of reproducibility in the field \cite{belz-etal-2021-systematic, belz-2022-metrological} and conducting reproducibility case studies \cite[e.g][]{coltekin-2020-verification, arvan-etal-2022-reproducibility-computational, fokkens-etal-2013-offspring}. However, compared to other fields relying on data collection and code production, in which publication venues have long-running dedicated reproducibility tracks \cite{sigir2022_reproducibility_track_2022, sigir2025_resource_reproducibility_2025}, code submission policies \cite{nature_portfolio_reporting_standards_2025, aaas_science_editorial_policies_2025, neurips_code_submission_policy_2025}, and official workflows in place to incentivize the sharing of (well documented and functioning) code and data \cite{sigmod_ari_2025, sigir_artifact_badging_2025}, the way reproducibility is handled in *CL venues still appears to be fragmented at best. 

\citet{mieskes-2017-quantitative} evaluates data availability of papers published at various *CL venues in 2016 and find that over 15\% of links to collected data provided in the papers were not available only a year after publication. They remark on the low share of data published on public hosting services such as GitHub, which they suggest might be more reliable due to their independence of personal or institutional webpages. \citet{wieling-etal-2018-squib} extend Mieskes' study and specifically analyse the availability of source code in papers published at ACL 2011 and 2016. They find that the availability of data and code as well as the share of working links had increased. In later work, \citet{mieskes-etal-2019-community} find GitHub to be the predominant source for code sharing in a community survey. Querying NLP researchers who attempted to replicate others' work, \citet{mieskes-etal-2019-community} and \citet{thomson-etal-2025-evolving} both identify resource and tool unavailability as common issues. \citet{arvan-etal-2022-reproducibility-computational} find a steady increase in papers with code published at *CL venues between 2016 and 2022, but also find that code availability alone does not ensure reproducibility. They were able to reproduce results for just two out of eight randomly selected papers. Several studies indicate that sharing research artifacts benefits authors, as code or data availability is associated with higher acceptance rates, improved reviewer scores, increased perceived reproducibility \cite{magnusson-etal-2023-reproducibility, chaudhuri_icml2019_code_submit_time_2019, pineau2021improving}, and higher citation counts \citep{KANG2023103477,colavizza2024analysis}. Our analysis complements these insights, by focusing specifically on the large-scale analysis of the availability of GitHub repositories linked in *CL venues over time.

\section{Methods}
\label{sec:methods}
We download all CL and ACL papers published between 2015 and 2025 using the ACL Anthology Python API.\footnote{\url{https://acl-anthology.readthedocs.io/py-v0.5.4/}} In order to determine whether trends in GitHub repository availability differ between conferences, we also download and parse all articles published at LREC and EACL 2024 as well as EMNLP, NAACL, AACL and COLING 2025 (including co-located events) to compare them with the ACL papers from 2025. 

\paragraph{Automatic Extraction and Availability Check} 
To extract GitHub links, we apply two strategies: we parse all embedded hyperlinks using the Python library \texttt{PyPDF2} and identify relevant links with a regex, and we additionally search the paper text for URLs containing \texttt{github.com} to capture links not embedded as hyperlinks, which is common in older papers. We then filter out GitHub links not pointing to a repository (e.g. links pointing to specific files or users), normalize the extracted links to the GitHub API format and query the GitHub API\footnote{\url{https://docs.github.com/en/rest?apiVersion=2022-11-28}} to determine repository availability and the number of files in each repository.\footnote{Automatic extraction of links was performed in June 2026.} 

\paragraph{Manual Verification and Categorization}
We manually verify all unique links flagged as unavailable or empty repositories by our script by opening the extracted URLs in a browser, clicking the links as rendered in the paper, and copying links directly from the paper, fixing formatting issues when necessary.\footnote{For example, line breaks sometimes introduce spurious characters such as ``\%20'' when copying links.} If none of these steps lead to a working repository, we search for the corresponding GitHub user and inspect their repositories to confirm that no repository with the reported name exists. We also manually inspect repositories containing only a single file to identify placeholder README files without substantial content. 

We define the following repository availability categories:
\begin{itemize}
\item \textbf{Available}: Repositories that return files via the GitHub API or are recoverable through manual verification and contain code or data.
\item \textbf{Unavailable}: Repositories that are inaccessible or lack substantive content. We further distinguish:\\
\textbf{404}: Links that return a 404 page and remain unrecoverable after manual verification; \\
\textbf{Empty}: Repositories without any files; \\
\textbf{Placeholder}: Repositories containing only a license or README without links to code or data or contact information, often limited to the paper title or a ``coming soon'' notice.
\end{itemize}

Finally, for all empty, unavailable, and placeholder repositories, we check the corresponding papers to determine whether the links referred to the paper’s own code or data or to related work. We conducted manual verification in June and July 2026.

\begin{figure*}[tb]
    \centering
    \begin{minipage}[t]{0.49\linewidth}
        \centering
        \includegraphics[width=0.8\linewidth]{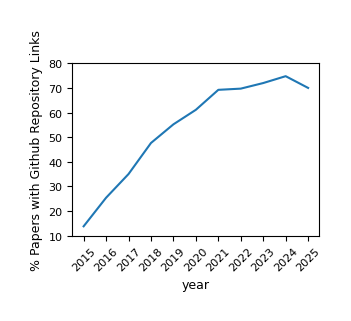}
        \caption{Share of ACL papers that point to GitHub}
        \label{fig:repo_development}
    \end{minipage}
    \hfill
    \begin{minipage}[t]{0.49\linewidth}
        \centering
        \includegraphics[width=\linewidth]{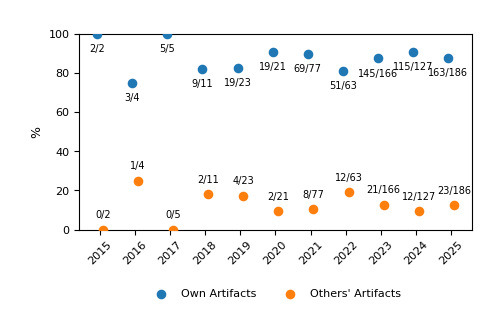}
       \caption{Shares of unavailable repositories pointing to ACL papers' own versus others' artifacts}
        \label{fig:own_repo_development}
    \end{minipage}
\end{figure*}
\section{Results}
\paragraph{Code Availability at ACL} We parse a total of 14,727 unique GitHub repository links from 12,087 ACL papers (our script does not return any links to GitHub repositories for 6,330 papers). 354 of the parsed repository links return a \textbf{404} and 326 are \textbf{empty} or contain only \textbf{placeholder} files. 88\% of links to unavailable repositories point to a papers' own research artifacts. Although references to unavailable artifacts from other works are less common, they appear even in recent publications, with rates remaining stable over time (see Figure \ref{fig:own_repo_development}). Figure \ref{fig:repo_development} shows a steady increase in the share of ACL papers linking to GitHub repositories over the years. Contrary to our expectations, the data also shows noticeable rates of unavailable repositories linked in recent years (see Figure \ref{fig:share_unavailable}).

\begin{figure}
    \centering
    \begin{minipage}[t]{0.49\linewidth}
        \centering
        \includegraphics[width=\linewidth]{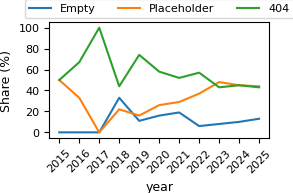}
        \caption{Development of unavailability types of GitHub links pointing to own artifacts over the years (ACL papers)}
        \label{fig:unavailability_types}
    \end{minipage}
    \hfill
    \begin{minipage}[t]{0.45\linewidth}
        \centering
        \includegraphics[width=\linewidth]{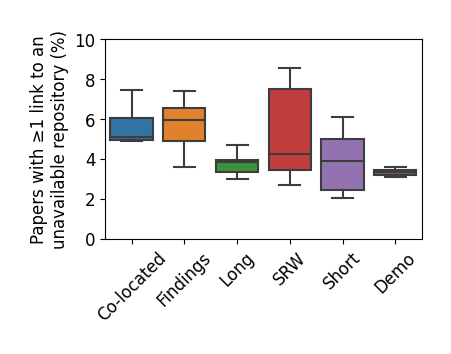}
        \caption{Annual shares of ACL papers linking to unavailable repositories by paper type/event 2021-2025}
        \label{fig:own_pie_right}
    \end{minipage}
\end{figure}

Upon closer inspection, one of the reasons for this is an increase in placeholder repositories: In 2025, 41\% of unavailable links consist solely of placeholder files, compared to a mean of 28\% in previous years. This trend is reversed for links leading to 404 errors, which made up 63\% of unavailable repositories on average in previous years and 47\% in 2025. A large share of these 404 errors accounts for papers' own code or data even in 2025 (80\%), which might indicate that these repositories have never been created (see Figure \ref{fig:unavailability_types}). Unavailable repositories most commonly appear in papers published at Findings of the ACL but also 
in the main track and at co-located events (see Figure \ref{fig:own_pie_right}).

\paragraph{Code Availability in Computational Linguistics} Out of 317 unique GitHub repositories extracted from the 369 papers published in CL between 2015 and 2025, only 7 (2\%) repositories are unavailable, six of which point to the paper's own code or data. Of these, four links lead to \textbf{404} errors, one (published in 2019) constitutes a \textbf{placeholder} repository, and one (published in 2022) is \textbf{empty}. The lower share of unavailable GitHub repositories compared to conferences suggests stronger curation and more stable code-sharing practices in journal publications, reflecting stricter editorial standards. It may also be influenced by the single-blind review format, which allows authors to include non-anonymous repository links at initial submission, increasing the likelihood that links are complete and functional at publication time. Still, this does not fully prevent unavailable repositories being linked. 

\subsection{Comparison with Other Conferences}
For comparison, we repeat the procedure outlined in §\ref{sec:methods} for six other computational linguistics conferences: LREC and EACL 2024 as well as NAACL, EMNLP, COLING, and AACL 2025. The share of papers linking to GitHub repositories ranges between 61\% (EACL) and 70\% (ACL) for all conferences. The general trends of unavailability do not differ much between conferences, as shown in Table \ref{tab:general_comp}. In most cases, between 4 and 5\% of the GitHub repositories found in articles were unavailable in 2024/2025. The exception to this is AACL, with a slightly higher share of unavailable links. EACL, AACL, and NAACL had a higher share of \textbf{404}'s and a lower share of \textbf{Placeholder} repositories than the other conferences, which could point to the fact that linked repositories had never existed or were never made public in the first place. On the other hand COLING has the lowest share of \textbf{empty} repositories, but a higher share of \textbf{Placeholder} repositories than the others. The relative similarity of overall shares and unavailability type distributions between conferences leads us to conclude that the observations made here are not ACL-specific, but pervasive, at least among computational linguistics conferences.
\begin{table}[t]
    \centering
    \resizebox{\textwidth}{!}{
    \begin{tabular}{l|r@{ }c@{ }l|c|c|c|c}
         \textbf{Conference}&\textbf{Unavailable} &/& \textbf{ Total Links (\%)}&\textbf{Own Code}& 
         \multicolumn{3}{c}{\textbf{Unavailability type}} \\
         & & & & &
         {\textbf{404}}&\textcolor{black}{\textbf{Empty}}&\textcolor{black}{\textbf{Placeholder}}\\
         \toprule
         \textbf{ACL}& 186 &/& 4454 (4\%)& 88\%&\textcolor{black}{43\%}&\textcolor{black}{13\%}&\textcolor{black}{44\%}\\
         \textbf{NAACL}& 85 &/& 1961 (4\%)& 87\%&\textcolor{black}{55\%}&\textcolor{black}{12\%}&\textcolor{black}{32\%}\\
         \textbf{EMNLP}&196 &/& 4104 (5\%)&92\%&\textcolor{black}{37\%}&\textcolor{black}{17\%}&\textcolor{black}{46\%}\\
         \textbf{EACL}& 37 &/& 826 (4\%)&95\%&\textcolor{black}{60\%}&\textcolor{black}{9\%}&\textcolor{black}{31\%}\\
         \textbf{AACL}& 38 &/& 551 (7\%)& 92\%&\textcolor{black}{51\%}&\textcolor{black}{17\%}&\textcolor{black}{31\%}\\
         \textbf{COLING}& 62 &/& 1219 (5\%)& 92\%&\textcolor{black}{42\%}&\textcolor{black}{7\%}&\textcolor{black}{51\%}\\
         \textbf{LREC}& 95 &/& 2188 (4\%) &94\%&\textcolor{black}{45\%}&\textcolor{black}{15\%}&\textcolor{black}{40\%}\\
         \bottomrule
    \end{tabular}
    }
    \caption{Share of papers with GitHub links at EACL and LREC 2024 as well as ACL, COLING, EMNLP, AACL and NAACL 2025 and percentage of those links pointing to unavailable repositories. Share of unavailable repositories for papers’ own code/data, with breakdown of unavailability types within that subset provided in brackets.}
    \label{tab:general_comp}
\end{table}

\subsection{Comparison with Other Online Repositories}
\begin{figure}[htbp]
    \centering
    \begin{minipage}[t]{0.49\linewidth}
        \centering
        \includegraphics[width=0.8\linewidth]{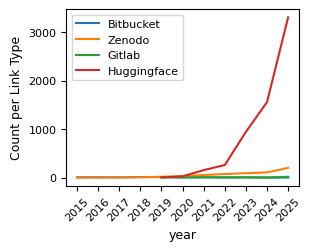}
        \caption{Occurrences of repository types other than GitHub in ACL papers over the past ten years}
        \label{fig:other_links}
    \end{minipage}
    \hfill
    \begin{minipage}[t]{0.49\linewidth}
        \centering
        \includegraphics[width=0.8\linewidth]{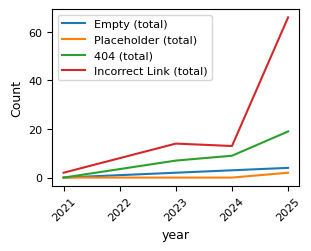}
        \caption{Occurrences of different unavailability types in unavailable HuggingFace repositories (ACL papers)}
        \label{fig:hf_unavailability}
    \end{minipage}
\end{figure}

We also parse links to other online repositories that can be used to store code and data, specifically GitLab, HuggingFace, and Bitbucket,
to evaluate trends in code-sharing platform usage. 
Additionally, we parse links to Zenodo, as it assigns DOIs and supports persistent, long-term storage of data and code.
While links to HuggingFace pages have become increasingly frequent in the past five years (see Figure \ref{fig:other_links}), Bitbucket, GitLab, and Zenodo are very rarely used, with only 58, 47, and 496 unique URLs found, respectively.

We check availability for Zenodo and HuggingFace links. We identify 141 unavailable HuggingFace repositories, corresponding to 3\% of all unique HuggingFace links. The majority of these (84\%) link to others' models or data, except for \textbf{Placeholder} (two overall, both pointing at own artifacts) and \textbf{Empty} repositories (six out of nine pointing to own artifacts). Manual verification shows most unavailable links are actually incorrect, meaning that the artifact exists but under a different path (e.g., renamed accounts or models), making it harder to identify the specific artifact referenced in the text (see Figure \ref{fig:hf_unavailability}). This suggests the main issue with HuggingFace link persistence is name changes over time, leading to deprecated links. This also suggests HuggingFace is used more as a source for existing models and data than as a platform for publishing new research artifacts.

Out of all unique Zenodo URLs, only two are unavailable, corresponding to 0.4\%, a much lower share than the 3\% unavailability rate for GitHub repositories linked in articles since 2015. This reinforces the notion that using persistent and safe code repositories facilitates long-term use of research outcomes and artifacts.

\section{Discussion and Recommendations}
Although making research artifacts available has gained increasing importance in NLP over the past years, we still find some failings in how this is handled in practice. We are particularly concerned in light of the share of papers that claim to open-source their code but end up not filling the repositories linked in the paper months or even years after publication or linking to repositories that do not exist. This may also be a consequence of growing pressures on the peer review process at major conferences. As submission numbers rise, reviewers may have less capacity to thoroughly verify repository contents. We see this as a breach of trust in the scientific publication process. Putting more emphasis on the validation of code availability and documentation in the peer review process would increase the potential for uptake of results and methods by other researchers \cite[see][inter alia]{KANG2023103477, pineau2021improving, chaudhuri_icml2019_code_submit_time_2019}. 
Based on our observations, we suggest a set of best practices for authors, reviewers, and publication chairs to decrease the share of newly published papers pointing to unavailable GitHub repositories.  \\
\textbf{For Authors:} 
\textit{Share code and data at time of paper submission}. Aiming to include code upon initial submission decreases the need for long code preparation times after acceptance or publication. This is strongly encouraged and sometimes even expected in other fields and has been shown to be beneficial for peer review \cite{chaudhuri_icml2019_code_submit_time_2019, neurips_code_submission_policy_2025, sigir2026_submission_policies_2025}.\footnote{The likelihood of authors achieving this aim may be higher if code and data submission deadlines are after the main peer review submission deadline, as implemented at  \citet{neurips_code_submission_policy_2025}, for example.} Whenever the identity of authors is not uncovered by the code or data itself, we recommend sharing it at the time of paper submission. This can, for instance, be done as a file upload to OpenReview, or by anonymizing existing repositories using \href{https://anonymous.4open.science}{Anonymous GitHub}, a free web service that fully anonymizes GitHub repositories and is frequently used for peer review in fields such as information retrieval \cite{sigir2026_submission_policies_2025, ictir2025_call_for_papers_2025} and recommended  for use in the ARR call for papers \cite{acl_arr_cfp}. 

\textit{Create persistent identifiers for code repositories}. Persistent identifiers are less volatile than repository links and will still resolve if, for example, repository or usernames are changed. This can be achieved by archiving repositories on Zenodo, for instance.\footnote{\url{https://docs.github.com/en/repositories/archiving-a-github-repository/referencing-and-citing-content}} For ARR, code and data can usually be included in the submission, and research artifacts generally also appear in the ACL Anthology. \\
\textbf{For Reviewers:} 
\textit{Be diligent about checking links and files provided during peer review}. If code or data are provided in a submission, reviewers should check links or submitted files at least briefly to make sure they are accessible and contents align with descriptions in the paper. Empty and placeholder repositories should be mentioned in the review.\\
\textbf{For Publication Chairs \& Editors:} 
\textit{Check camera-ready papers for broken links and invalid GitHub references}. Camera-ready versions should be checked for faulty links and unavailable GitHub repositories. To facilitate this, we provide code based on our extraction script integrated in the aclpubcheck workflow currently used for publication at *CL conferences and workshops with this paper.

To evaluate the tool's practical feasibility, we randomly sampled 50 ACL papers flagged as containing potentially problematic repository links, stratified by publication year. As a control, we sampled 50 unflagged papers from the same venues and years. An independent evaluator manually verified all referenced GitHub repositories using only the PDFs. The tool identified papers with problematic links with 89\% accuracy and 78\% precision. Eleven papers were false positives, for which a repository exists, but URLs contained, for example, spelling or hyperlinking errors. No unavailable repositories were found in the control sample, suggesting high recall.\footnote{Because unavailable repositories are relatively rare, the absence of false negatives in our sample should not be interpreted as evidence of perfect recall. Some problematic 
links may remain undetected.} Based on the manual verification of all links flagged as problematic (see §\ref{sec:methods}), the tool achieved 62\% precision on the paper level, improving from 25\% (2015) to 72\% (2025), largely due to better PDF hyperlinking. For the 2025 proceedings, this would mean that most workshop chairs would have to review only 1–5 papers, while main conference chairs would have to review 84 long, 95 findings, and 5 short papers. By revealing automatic linking problems, even false positives flagged by the tool can likely improve editorial quality and access to resources.

\section{Conclusion}
In light of increased code and data sharing at *CL venues, this work focuses on the (long-term) availability of online repositories linked in papers published at CL and ACL in the past 10 years. We find a concerning persistency of unavailability rates and placeholder repositories which are not filled with code after publication. With publication counts at *CL venues rising, these rates translate into an accelerating growth of unavailable repositories and artifacts year after year.
Not making code and data available significantly mitigates the reusability of methods introduced and trustworthiness of results. We outline recommendations to support reliable access to resources over time and expand the aclpubcheck library with a lightweight method to check papers for invalid or broken GitHub references. We encourage authors and publication chairs to introduce this method into the current publication workflow to ensure the availability of linked research artifacts and the usefulness of research published at *CL venues in the future. 

While we see code and data sharing as a basic requirement for open science, this does not ensure reproducibility of reported results \citep{arvan-etal-2022-reproducibility-computational}. Moreover, this work does not assess whether papers that should share code or data actually do so. Instead, we focus on the long-term availability of repositories, often linked from papers that claim to share research artifacts. In future work, we plan to expand on the findings presented here by focusing on ways to automatically evaluate the quality and documentation of provided code, allowing us to draw insights on reproducibility in addition to code availability.

\section{Acknowledgements}
Codex and ChatGPT were used for coding assistance and code documentation. The authors take full responsibility for all reported results.

\bibliographystyle{compling}
\bibliography{COLI_template}

\end{document}